\documentclass{article}
\usepackage{spconf, amsmath, graphicx}
\usepackage{amssymb}
\usepackage{tcolorbox}

\newcommand{\m}[1]{\mathbf{#1}}
\newcommand{\p}[1]{\partial{#1}}
\newcommand{\norm}[1]{\left\lVert#1\right\rVert}
\newcommand{\BigO}[1]{\ensuremath{\operatorname{O}\bigl(#1\bigr)}}

\title{Complex Evolution Recurrent Neural Networks (\lowercase{ce}RNN\lowercase{s})}

\name{Izhak Shafran, Tom Bagby, and R.~J.~Skerry-Ryan}
\address{Google Inc}

\begin{document}
\maketitle

\begin{abstract}
Unitary Evolution Recurrent Neural Networks (uRNNs) have three attractive properties: (a) the unitary property, (b) the complex-valued nature, and (c) their efficient linear operators~\cite{Arjovsky16}. The literature so far does not address -- how critical is the unitary property of the model? Furthermore, uRNNs have not been evaluated on large tasks. To study these shortcomings, we propose the complex evolution Recurrent Neural Networks (ceRNNs), which is similar to uRNNs but drops the unitary property selectively. On a simple multivariate linear regression task, we illustrate that dropping the constraints improves the learning trajectory. In copy memory task, ceRNNs and uRNNs perform identically, demonstrating that their superior performance over LSTMs is due to complex-valued nature and their linear operators. In a large scale real-world speech recognition, we find that pre-pending a uRNN degrades the performance of our baseline LSTM acoustic models, while pre-pending a ceRNN improves the performance over the baseline by 0.8\% absolute WER.  
\end{abstract}


\section{Unitary Evolution Recurrent Neural Networks (\lowercase{u}RNNs)}
\label{sec:uRNN}

Unitary evolution RNNs (uRNNs) were proposed to address the exploding gradients in RNNs, essentially by constraining the Eigenvalues of the linear transformations to be unity. We highlight the fundamentals of this model briefly to provide the context for our work. To understand why the gradients do not explode in this model, consider the standard RNN equation. 
\begin{eqnarray}
z_{t+1} &=& \m{W}_t h_t + \m{V}_t x_{t+1} \\
h_{t+1} &=& \sigma(z_{t+1}) 
\end{eqnarray}

For $T$ unrolled steps and cost $C$ at $T$, the gradients at any previous $t$ are:
\begin{eqnarray}
\frac{\p{C}}{\p{h_t}} &=& \frac{\p{C}}{\p{h_T}}  \frac{\p{h_T}}{\p{h_t}} 
                                  = \frac{\p{C}}{\p{h_T}} \prod_{k=t}^{T-1}  \frac{\p{h_{k+1}}}{\p{h_k}} \nonumber \\
                                &=& \frac{\p{C}}{\p{h_T}} \prod_{k=t}^{T-1} \m{D}_{k+1} \m{W}_k^T
\end{eqnarray}
where $\m{D}_{k} = $ diag$(\sigma^\prime(z_{k}))$ is the Jacobian of point-wise non-linearity. By imposing the constraint  $\norm{W_k}=1$ and using ReLu ($\sigma$) as the non-linearity activation function, the uRNNs guarantee that the upper bound will not explode/vanish. As an aside, while the upper bound will not explode or vanish, the gradients themselves may decrease rapidly.
\begin{eqnarray}
\norm{\frac{\p{C}}{\p{h_t}}} \leq \norm{\frac{\p{C}}{\p{h_T}}} \prod_{k=1}^{T-1} \m{D}_{k+1} = \norm{\frac{\p{C}}{\p{h_T}}}
\end{eqnarray}

In uRNNs, the transition matrix $W: x \mapsto y, \; \; x, y \in \mathbb{R}^N$ was replaced by a cascade of efficient {\bf unitary} linear operators, specifically Fourier transformation ($\cal F$), its inverse (${\cal F}^{-1}$), unitary diagonal matrix multiplication $U$, column permutation operator $P$, and Householder reflection ($R=\mathbf{I} - 2 \frac{v v^*}{||v||}, \; \forall v \in \mathbb{C}^n$). The operators -- $\{ {\cal F}, {\cal F}^{-1}, R, P\}$ -- are unitary in nature since its inverse is its conjugate transpose. The diagonal matrix is constrained to be unitary by choosing the elements on the unit circle, $U_{i,i} = e^{-i\theta_i }, \; \forall \theta_i \in \Re$. The cascade of operators were strung together as in Equation~\ref{eqn:unitary}.
\begin{eqnarray}
z = W_u x = U_3 R_2 {\cal F}^{-1} U_2 P R_1 {\cal F}_1 U_1 x \label{eqn:unitary}
\end{eqnarray}
Since the output of this transformation is a complex-valued vector, a complex-valued non-linear operator was needed. For this purpose, they use modReLu, a complex-valued extension of the ReLu, parameterized by a bias term $b$.
\begin{eqnarray}
\sigma_{modReLu}(z) = \sigma_{ReLu}(||z|| + b) \frac{z}{||z||}, \; \; z \in \mathbb{C} 
\end{eqnarray}
The resulting models has been found to converge faster than LSTMs in copy memory and addition tasks. The performance on pixel-by-pixel MNIST was mixed, the uRNNs outperformed LSTMs in the permuted case, but not in the un-permuted case. Their models have not been applied to large scale tasks such as automatic speech recognition, which we investigate in this work.

\section{Complex Evolution Recurrent Neural Networks (\lowercase{ce}RNNs)}

As mentioned in the abstract, the uRNN models have three attractive properties: (a) the unitary property, (b) the complex-valued nature, and (c) efficient linear operators. One of the questions that has not been addressed in the literature so far is how critical is the unitary property of the model. Clearly, the copy-memory task is designed to test the capacity of the model to retain the memory uncorrupted for long durations. The non-decaying property of the unitary transforms makes it particularly well-suited to address that task. However, in many real-world tasks such as automatic speech recognition (ASR) the model is only required to retain short-term memory and the ability to decay or forget may be useful. For answering these question, we create a variant that removes the unitary constraints selectively. We drop the unitary constraint on the diagonal matrices $D$ as in Equation~\ref{eqn:wbet}. 
\begin{eqnarray}
z = W_{ce} x = D_3 R_2 {\cal F}^{-1} D_2 P R_1 {\cal F}_1 D_1 x \label{eqn:wbet}
\end{eqnarray}
While any cascade of the unitary operators will result in a unitary linear transformation, the uRNN model chooses a specific cascade of transformation, as in Equation~\ref{eqn:unitary}, without any justification, but presumably based on empirical  evaluations. As such, the uRNN contains only $7N$ real-valued parameters and has limited model capacity since the resulting matrices do not span the entire space of a unconstrained matrix with $N^2$ parameters. This was one of the criticisms that inspired the "full unitary" model~\cite{Wisdom16}. On closer examination, this is not an inherent weakness and the model capacity can be increased arbitrarily by chaining more transformations in the cascade. For the purpose of head to head comparison, we retain the same cascade here.

We refer to the family of models that are constructed by chaining a cascade of the efficient complex-valued linear operators and non-unitary diagonal matrices as complex evolution RNNs (ceRNNs). The unconstrained diagonal could potentially take on the unitary values and hence the ceRNNs can be viewed as a superset of uRNNs. The complex-valued layers by themselves have several useful properties. For example, the orthogonal properties of their decision boundaries helps solve the XOR problem with fewer nodes~\cite{Nitta:2009, Hirose:2006} and they have better generalization characteristics~\cite{Hirose:2012}.

\section{Understanding the \lowercase{ce}RNN\lowercase{s}}
\subsection{Multivariate Linear Regression Task}


Consider a multivariate linear process, $y = W_m x + n, \; x, y, n \in \Re^N$, where $n$ denotes uncorrelated and independent noise, and $W_m$ is a randomly generated matrix that we wish to estimate using complex evolution matrix with a mean squared error (MSE) cost function. We compare the MSE cost and the convergence under three different settings -- (a) a matrix, $W$, (b) the unitary model defined in Equation~\ref{eqn:unitary}, and (c) a complex evolution model defined in Equation~\ref{eqn:wbet}. Note, while this is a convex problem, the matrices have different constraints and as a result they are not guaranteed to reach the same value of MSE loss at convergence. 

\begin{figure}[h]
\centering
\includegraphics[width=0.8\columnwidth]{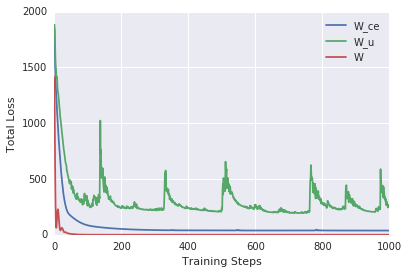}
\caption{Comparison of MSE with complex evolution (W\_ce), unitary (W\_u) and full matrix (W) models for the multivariate regression task.}
\label{fig:linreg_total_loss}
\end{figure}

\begin{figure}[h]
\centering
\includegraphics[width=0.8\columnwidth]{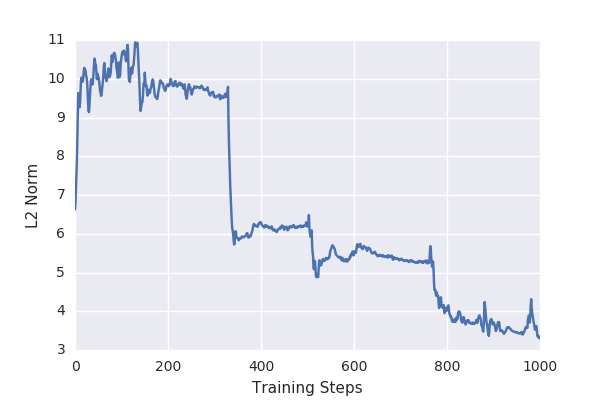}
\caption{Norm of the angles (radians) in the diagonal ($U_1$) as the training progresses.}
\label{fig:urnn_diag1}
\end{figure}

Figure~\ref{fig:linreg_total_loss} shows the convergence of the loss function for the three models. At convergence, the MSE is the least for the full matrix model (MSE=0.5) as expected. The unitary model performs poorly (MSE $\approx$ 250) even with more iterations than shown in the plot. When the learning rates is increased, we noticed spikes with loss more than 2k! To understand these spikes, we examined the L2 norm of the angles (in radians) of the elements of the diagonal matrices. Interestingly, as shown in Figure~\ref{fig:urnn_diag1}, the norm exhibits abrupt transitions that coincide with the spikes in the MSE loss function.


\begin{figure}[h]
\centering
\includegraphics[width=0.7\columnwidth]{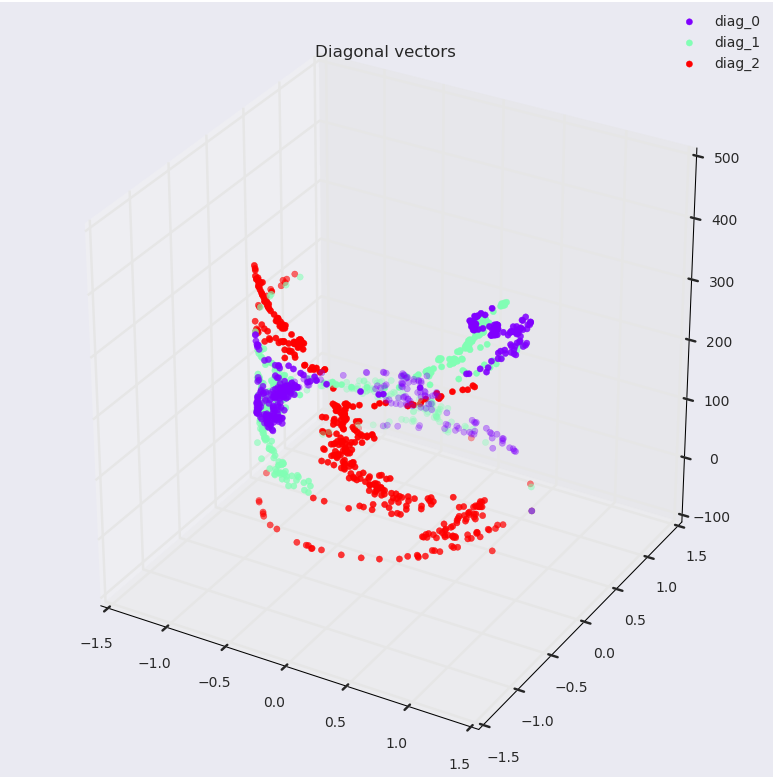}
\caption{The estimated diagonal components, diag\_i, in the unitary transform for the linear regression task where the x-y plane is the complex plane.}
\label{fig:diagonal}
\end{figure}

The complex evolution model on the other hand behaves better, with a smoother convergence and lower cost function (MSE=38) than the unitary model. We did not notice any of the erratic spiking behavior. In unitary matrix, as illustrated in Figure~\ref{fig:diagonal},  the diagonal components $(U_{i,i})$, lie on unit circle. In complex evolution model, this is not the case. 

\subsection{Copy Memory Task}
Next, we examine the performance on the copy memory task~\cite{Arjovsky16}, which is often employed to evaluate the memory capacity of the model. We compare the performance of three types of memory networks: (a) a long short term memory model (LSTM)~\cite{Hochreiter97}, (b) a unitary RNN with $W_u$ state-evolution transformation, and (c) an RNN with $W_{ce}$ state-evolution transformation. All three models were trained with cross-entropy cost function using Adam.

The input in a copy memory task consists of four components: (1) a target sequence ($T$), say of length 10, whose elements are sampled with repetition from a set of symbols $\{a_{i \in \{1:7\}}\}$, (2) an $N$-length filler sequence ($F_N$) with filler symbol $a_8$, (3) a trigger ($R$) with symbol $a_9$, and (4) another filler sequence ($F_T$), of the same length as $T$, with filler symbol $a_8$. The model is expected to output a sequence of filler symbols $a_8$ till it encounters the trigger symbol $R$ in the input and then it is expected to regurgitate the memorized target sequence $T$ provided in the first component of the input. 
 
\begin{figure}[h]
\centering
\includegraphics[width=0.8\columnwidth]{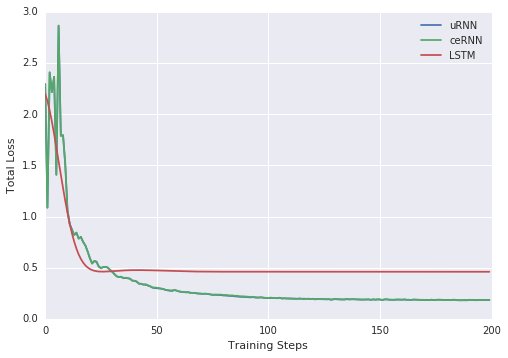}
\caption{Comparison of CE loss with the ceRNN, the uRNN and the LSTM for the copy memory task. ceRNN and uRNN are identical.}
\label{fig:copytask_loss}
\end{figure}

The cross-entropy loss of all the three models are shown in Figure~\ref{fig:copytask_loss}. The uRNN and ceRNN models have identical plots and their cross-entropy loss (0.18) are significantly better than the LSTM (0.46). The identical performance of uRNN and ceRNN models is not entirely surprising since the ceRNNs retain 3 of the 4 unitary components -- Fourier transform, its inverse and permutation transform.

\section{Automatic Speech Recognition}

The training data consists of about $15000$ hours of spontaneous speech from anonymized, human-transcribed Google Voice Search queries. For noise robustness training, each utterance in the training data is corrupted with additive noise and reverberation, corresponding to $100$ different settings of room size and signal-to-noise ratios. The performance of the models is evaluated on two test sets: an uncorrupted or \textit{clean} and corrupted or \textit{noisy} (in the same fashion as the training data) version of a 25-hour subset of utterances from Google voice search traffic. Each test set is about $25$ hours. The performance is measured against human transcriptions.  

The acoustic features consist of a sequence of vectors, computed at a rate of one vector per 30ms. Each vector is comprised of four stacked frames; the current frame along with three previous ones. Each frame contains 128 log mel-warped filter-bank outputs, computed at a rate of 100 frames per second using a sliding window of 32ms~\cite{Pundak16}.  During training, to emulate the effect of the right context, we delay the output state label by 5 frames~\cite{Sak14}. Each frame is labeled with one out of 8192 context-dependent output phoneme states. For more details about the model, see~\cite{Bagby17}. We adopt their LSTM baseline with 5-layers and 768 nodes in each layer.

We investigate inserting uRNN and ceRNN layers between the inputs and the LSTM baseline. Note, our preliminary experiments with uRNN and ceRNN layers by themselves performed poorly compared to LSTMs so we did not explore that any further. The poor performance is likely because both uRNN and ceRNN lack the sort of gating mechanism present in LSTMs. 

The uRNN+LSTM and ceRNN+LSTM models were trained from scratch to reduce cross-entropy using the Adam optimizer (\cite{KingmaB14}) with a batch size of 256. Attempts to train a simple RNN as a test layer with different learning rate failed. The results in the Table~\ref{table:wer1} are reported at convergence (50M training steps). The unitary RNN layer hurts the ASR performance when compared to the baseline system. On the other hand, the ceRNN gives a marginal and consistent gain over the baseline system. To rule out the possibility that the ceRNN is only renormalizing the data, we introduced a normalization layer before uRNN. That still does not help improve the performance of the uRNN models.

\begin{table}[h]
\centering
\begin{tabular}{|c|c|c|}
\hline
Model & Clean & Noisy \\ \hline \hline
LSTM & 11.7 & 18.5 \\
uRNN (512) + LSTM & 12.6 & 20.9 \\
ceRNN (512) + LSTM & 11.5 & 18.3 \\ \hline
\end{tabular}
\caption{Performance (WER) comparison of the LSTM, the uRNN+LSTM, and the ceRNN+LSTM models.}
\label{table:wer1}
\end{table} 

Since the ceRNN model parameters grow only linearly in the size (nodes) of the layer in contrast to standard matrix-based layers ($\BigO{7N} vs.~\BigO{N^2}$), the computational cost also scales linearly. This makes it practical to increase the size of the layer from 512 to 1024 and 2048 and Table~\ref{table:wer2} reports the associated improvements we observed. The ceRNN with 2048 nodes (with only 14k additional parameters) appears to provide the most performance gain. Increasing the number of nodes further to 4096 increased the performance only marginally. 

\begin{table}[h]
\centering
\begin{tabular}{|c|c|c|c|}
\hline
Model & Clean & Noisy \\ \hline \hline
ceRNN (512) + LSTM & 11.5 & 18.3 \\
ceRNN (1024) + LSTM & 11.4 & 18.1 \\ 
ceRNN (2048) + LSTM & 11.1 & 17.5 \\ \hline
\end{tabular}
\caption{Performance (WER) improvements observed with 512, 1024 and 2048 ceRNN nodes.}
\label{table:wer2}
\end{table} 

Increasing the ceRNN output also increases the size of the LSTM inputs and the matrices associated with the gates. To rule out the possibility that the performance improvements are related to the larger size of the LSTM gates incidentally caused by using ceRNN, we ran a contrasting experiment where only 512 outputs from the ceRNN were retained. The rest of the outputs were discarded. This is analogous to zero-padding the FFT and demonstrates how the ceRNN can be applied to project from higher to lower dimensions without an explicit linear bottleneck layer. In our experiments, we found that the performance does not suffer from discarding the extra dimensions and retaining the same number of parameters on the LSTM gates. Thus, the improvements are from the ceRNN layer itself.  

\begin{figure}[h]
\centerline{\includegraphics[width=0.9\columnwidth]{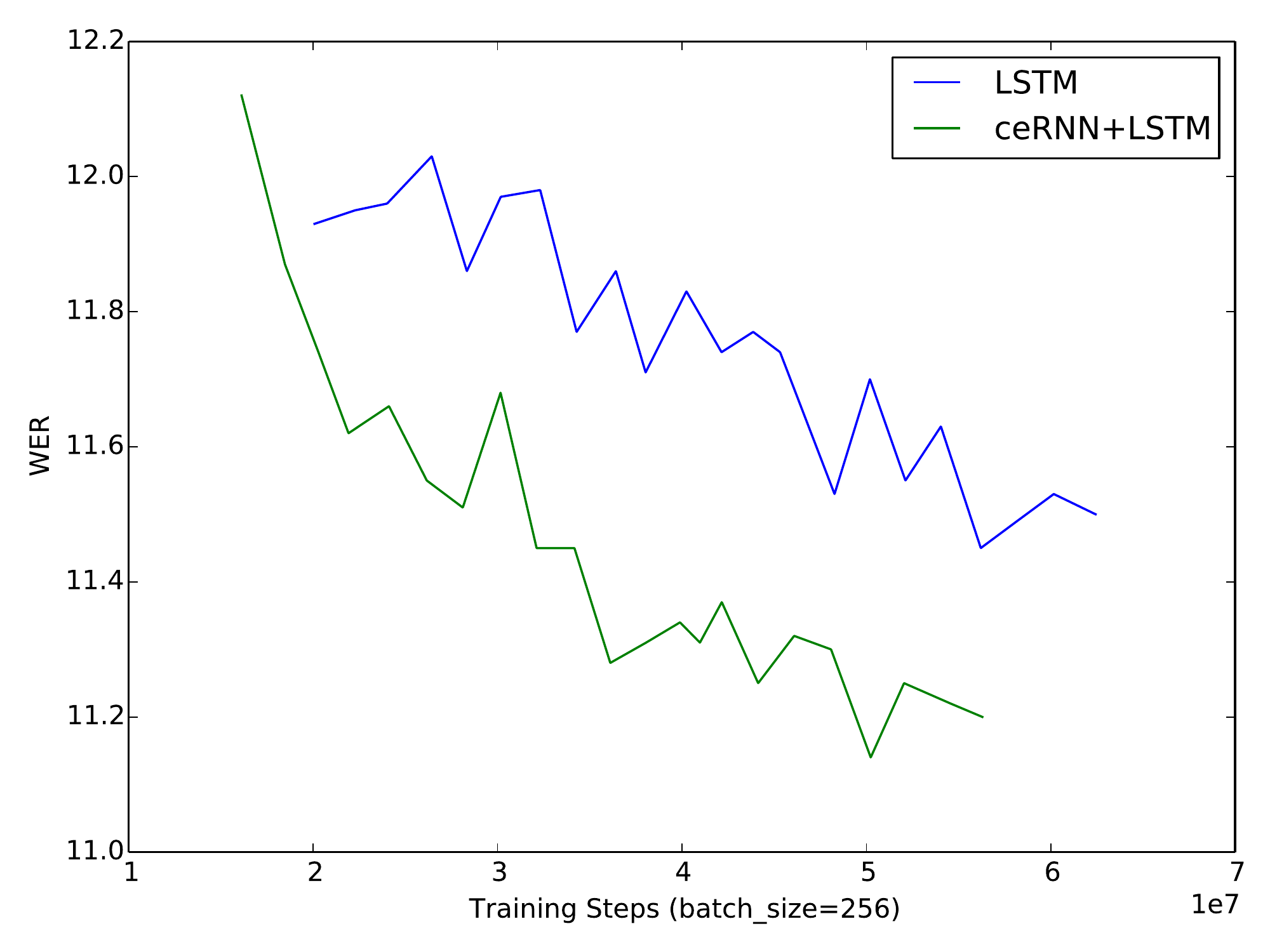}} 
\centerline{\includegraphics[width=0.9\columnwidth]{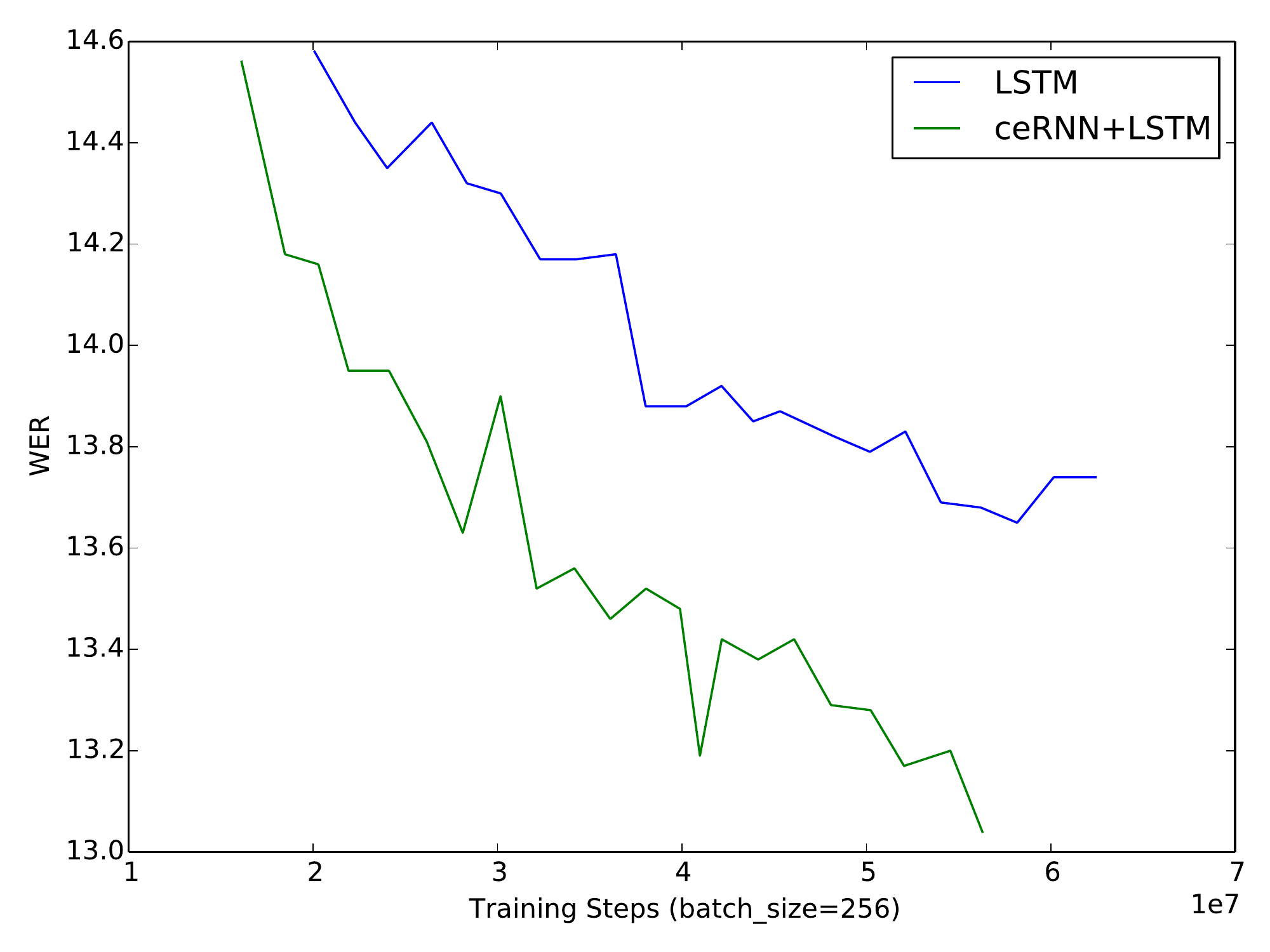}}
\caption{Performance of the ceRNN+LSTM model as training progresses: (a) on clean, and (b) noisy test data.} \label{fig:werplot}
\end{figure}

In Figure~\ref{fig:werplot}, we compare the performance of the baseline LSTM model with the best ceRNN+LSTM model as a function of training steps. The ceRNN+LSTM model is consistently better than the LSTM models for all the checkpoints on both the data sets, improving the performance by about 0.5\% absolute WER over the LSTM models at convergence. 

\section{Conclusions and Future Work}
In this work, we propose ceRNNs, a family of RNNs whose transition matrix consists of a cascade of efficient complex-valued (unitary) linear operators and non-unitary diagonal matrices. We illustrate that dropping the unitary constraint on the diagonal improves the learning trajectory of a simple multivariate regression model. On copy memory task, we find ceRNNs and uRNNs perform identically, demonstrating that strict unitary constraint is not necessary to retain memory. In a head to head comparison of ceRNN and uRNN layer prepended before our baseline LSTM acoustic model, we find that uRNNs degrade the performance while ceRNNs improves the performance over our baseline, achieving as much as 0.8\% absolute WER improvement. In our experiments, we also demonstrate that these matrices can model non-square linear transformations by dropping the extra dimensions, thus avoiding the expense of a bottleneck layer. Since these models utilize complex-valued FFTs as inputs, they open up the possibility of integrating de-reverberation and beam-forming within this framework. The linear transforms can be stitched together in many different ways to create a family of models with different trade-offs in memory and computational efficiency. This work raises the question whether the gains observed in different flavors of unitary matrices are due to their unitary nature or the properties (e.g., regularization) of transformations use to create those matrices~\cite{Jing17,Mhammedi17}. 

\section*{Acknowledgements}
Thanks to Matt Shannon for useful discussions on the properties of unitary matrices and the bounds.

\bibliographystyle{IEEEbib}
\bibliography{ceRNNs}

\end{document}